\documentclass[11pt]{article}

\usepackage[preprint]{acl}

\usepackage{times}
\usepackage{latexsym}

\usepackage[T1]{fontenc}

\usepackage[utf8]{inputenc}

\usepackage{microtype}

\usepackage{inconsolata}

\usepackage{graphicx}

\usepackage{enumitem}
\usepackage{booktabs}
\usepackage{multirow}
\usepackage{amsmath}
\usepackage{amssymb}
\usepackage{array}
\usepackage[table]{xcolor}
\usepackage{threeparttable}
\usepackage{lipsum}
\usepackage{xspace}

\newcommand{\model}{MolGram\xspace} %

\title{Augmenting Molecular Language Models with Local $n$-gram Memory}

\newcommand{\aff}[1]{\textsuperscript{#1}}

\author{%
    \begingroup
    \renewcommand{\arraystretch}{1.02}
    \begin{tabular}{c}
    \textbf{Xinni Zhang\aff{1} \quad
    Zijing Liu\aff{2}\quad
    He Cao\aff{2} \quad
    Yu Li\aff{2}\quad 
    Irwin King\aff{1}} \\[0.5em]
    \normalfont \aff{1}The Chinese University of Hong Kong \quad
    \normalfont \aff{2}International Digital Economy Academy \\
    \end{tabular}
    \endgroup
}


\begin{document}
\maketitle

\begin{abstract}
Transformer-based language models for SMILES strings suffer from a locality gap: standard character-level tokenization fragments chemically meaningful motifs, forcing models to repeatedly learn local syntax at the expense of long-range dependencies.
To address this without disrupting standard tokenizers, we propose \model, which integrates a conditional $n$-gram memory module into molecular language models. \model maps local string patterns to learned embeddings via scalable hash lookups and dynamically injects this regional context into hidden states. Evaluations across three tasks, including unconditional molecule generation, forward reaction prediction, and single-step retrosynthesis, show that \model consistently improves performance.
Crucially, our analyses demonstrate that \model outperforms baselines with 3$\times$ more parameters, establishing explicit local pattern memory as a highly efficient inductive bias.
\end{abstract}

\section{Introduction}
\label{sec:intro}

SMILES~\citep{weininger1988smiles} represents molecules as character sequences, making Transformer-based language models a natural fit. These models have achieved strong results in reaction prediction~\citep{schwaller2019molecular,sagawa2023reactiont5}, retrosynthesis planning~\citep{tetko2020state,deng2025rsgpt}, and molecular generation~\citep{bagal2021molgpt}. 
However, the standard character-level tokenization fragments chemically meaningful motifs across multiple tokens: a carbonyl group \texttt{C(=O)} uses five tokens, and a benzene ring \texttt{c1ccccc1} has eight. Consequently, the model must rediscover these recurring substructures via the language-model objective. We refer to this mismatch as the \emph{locality gap}: model capacity that should support global reaction logic and long-range dependencies is instead spent relearning local chemistry across multiple positions and layers.

Prior work has mainly addressed this gap by changing the tokenizer or the data, for example, utilizing BPE~\citep{sennrich2016neural}, fragment-based representations such as SAFE~\citep{noutahi2024gotta} and fragSMILES~\citep{mastrolorito2025fragsmiles}, or SMILES augmentation~\citep{bjerrum2017smiles}. 
While effective, these approaches often complicate downstream data processing and disrupt the universal, standardized syntax of SMILES strings.

In this work, we take a model-side approach and propose \model, which is a \textbf{Mol}ecular language model equipped with En\textbf{gram}~\citep{cheng2026conditional}, a conditional n-gram memory module. \model maps local n-gram patterns to learned embeddings via hash-based lookup and injects them into hidden states through a learned gate, while keeping the original character-level representation unchanged. This is a particularly good fit for SMILES, where short contiguous n-grams often align with recurring chemical motifs such as functional groups, bond patterns, and ring closures.
We evaluate \model on three tasks: unconditional molecule generation~\citep{brown2019guacamol,polykovskiy2020molecular}, forward reaction prediction~\citep{schneider2016s}, and single-step retrosynthesis~\citep{schneider2016s}, with both decoder-only and encoder-decoder architectures~\citep{radford2019language,raffel2019exploring}. Across all tasks, \model consistently improves both generation and reaction prediction.
Furthermore, the improvements are not due to extra capacity: larger baselines with up to 3$\times$ more parameters still cannot match \model.

Our contributions are threefold: (1) we propose \model, the first n-gram memory-augmented molecular language model, with architecture-specific adaptation for MolGPT and T5Chem backbones; (2) comprehensive evaluation across diverse chemical tasks; and (3) a rigorous analysis showing that the gains come from explicit local pattern memory rather than increased capacity.

\section{Related Work}
\label{sec:related}

Transformer-based chemical language models have become a standard choice for SMILES-based molecular modeling since the Molecular Transformer~\citep{schwaller2019molecular}. Subsequent work has explored decoder-only models (MolGPT;~\citealt{bagal2021molgpt}), encoder-decoder architectures (T5Chem,~\citealt{lu2022unified}; Chemformer,~\citealt{irwin2022chemformer}; ReactionT5,~\citealt{sagawa2023reactiont5}), and large-scale pretraining (ChemBERTa,~\citealt{chithrananda2020chemberta}; MoLFormer,~\citealt{ross2022large}; RSGPT,~\citealt{deng2025rsgpt}), as well as recent LLM-based multi-task learning and RL post-training for retrosynthesis~\citep{lin2025enhancing,zhang2025reasoning}. To address SMILES variability, prior work has mainly relied on tokenizer- or data-level strategies, including fragment-based tokenization (SAFE,~\citealt{noutahi2024gotta}; fragSMILES,~\citealt{mastrolorito2025fragsmiles}) and SMILES enumeration~\citep{bjerrum2017smiles,tetko2020state}. In contrast, explicit local-pattern memory offers a model-side alternative: n-gram-based sparse memory has evolved from classical methods to architectures such as Engram~\citep{cheng2026conditional} and Gengram~\citep{xu2026beyond}. Building on this line of work, we apply Engram to SMILES while retaining simple character tokenization and evaluate it with GPT and T5 backbones.

\section{Method}
\label{sec:method}

\begin{figure}[t]
    \centering
    \resizebox{1\linewidth}{!}{
    \includegraphics{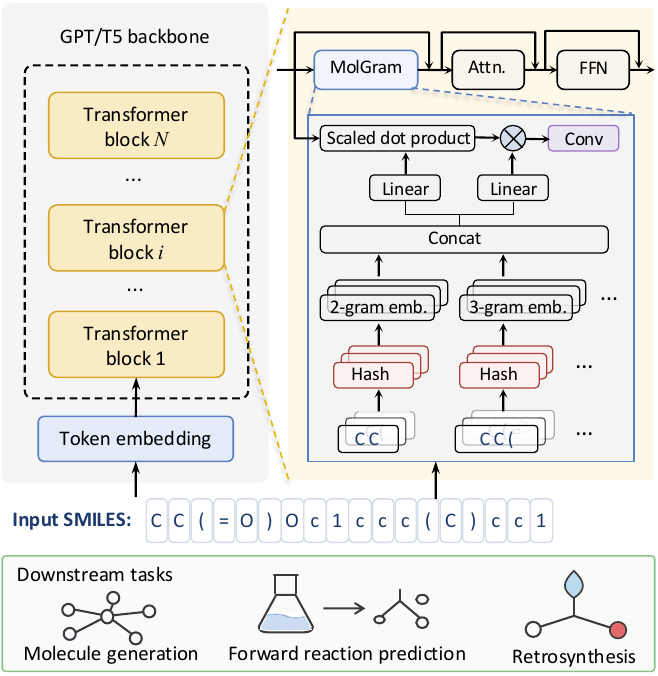}}
    \vspace{-0.2cm}
    \caption{Architecture overview of \model. At selected Transformer layers, the module hashes local n-grams into a sparse embedding table, gates the retrieved representations against the backbone hidden state, and applies a short convolution for local composition.}
    \label{fig:model}
    \vspace{-0.4cm}
\end{figure}

We augment Transformer-based molecular language models with \model, a lightweight conditional memory module that explicitly captures local n-gram patterns in SMILES sequences.
In character-level SMILES, short contiguous token windows often correspond to recurring chemical substructures such as functional groups, ring systems, and bond motifs.
Rather than relying on the Transformer to rediscover these patterns implicitly through attention, \model provides a dedicated lookup mechanism that stores and retrieves learned representations of such patterns, injecting them into selected Transformer layers as illustrated in Figure~\ref{fig:model}.

\subsection{\model Module}

The \model module is inserted at selected Transformer layers and operates on the hidden representations produced by the backbone. Given the hidden state $\mathbf{h}_t^{(i)} \in \mathbb{R}^{d}$ at position $t$ of layer $i$ and the token sequence $\mathbf{x} = (x_1, x_2, \ldots, x_T)$ from the tokenized SMILES input, the module produces a local-pattern enriched representation $\hat{\mathbf{h}}_t^{(i)}$ that replaces $\mathbf{h}_t^{(i)}$ before passing to subsequent layers. It consists of three components: n-gram hash mapping, substructure-aware gating, and short convolution (Figure~\ref{fig:model}).\footnote{For notational clarity, we omit the layer index superscript $(i)$ in the remainder of this section; all operations described below are applied identically at each insertion layer.}

\paragraph{N-gram hash mapping.}
The goal of this component is to build a dictionary-like memory that maps recurring local token patterns to learned representations. We implement this via hashing: each n-gram is mapped to an index in a large embedding table, enabling efficient $O(1)$ lookup without storing explicit n-gram strings.

Given the token sequence $\mathbf{x}$, for each position $t$ and each n-gram size $n$ ($n_{\text{min}} \leq n \leq n_{\text{max}}$), we first compute a \emph{mixed key} that combines the $n$ token IDs into a single integer:

\begin{equation}
    \text{key}_t^{(n)} = \bigoplus_{j=0}^{n-1}\, m_j \cdot x_{t-j},
    \label{eq:calc_key}
\end{equation}
where $m_j$ are pre-generated large odd-integer multipliers that ensure distinct n-grams produce well-separated keys, and $\bigoplus$ denotes the bitwise XOR operation. The XOR-based mixing avoids the information loss inherent in simple addition while remaining computationally inexpensive.

The mixed key is then mapped to a table index via modular hashing:
\begin{equation}
    b_t^{(n)} = \text{key}_t^{(n)} \bmod S^{(n)},
    \label{eq:hash}
\end{equation}
where $S^{(n)}$ is the hash table size for n-gram size $n$, set to the smallest prime no less than a user-specified capacity; this choice yields a more uniform distribution of entries across the table.

To further mitigate collisions, we adopt a \textit{multi-head} design: the table is partitioned into $M$ independent heads, each with its own prime modulus $S_m^{(n)}$. For each head $m$, a separate hash index $b_{t,m}^{(n)}$ is computed via Eq.~\ref{eq:hash}, and the corresponding embedding $\mathbf{e}_{t,m}^{(n)} \in \mathbb{R}^{d_h}$ is retrieved. The embeddings from all heads and all n-gram sizes are concatenated:
\begin{equation}
    \mathbf{e}_t = \Big\Vert_{n=n_{\text{min}}}^{n_{\text{max}}} \Big\Vert_{m=1}^{M} \mathbf{e}_{t,m}^{(n)},
    \label{eq:embed}
\end{equation}
where $\mathbf{e}_t \in \mathbb{R}^{M(n_{\text{max}}-n_{\text{min}}+1)\,d_h}$, allowing the module to capture diverse aspects of each pattern.

\paragraph{Substructure-aware gating.}
The embeddings retrieved from the hash table are \emph{static}: they encode general n-gram pattern information but lack awareness of the current molecular context. To selectively retain informative substructure signals and suppress irrelevant retrievals, we introduce a gating mechanism that leverages the backbone's hidden state to assess the relevance of each retrieved embedding and modulate its contribution accordingly.

We first project the concatenated n-gram embedding $\mathbf{e}_t$ into a key and a value vector:
\begin{equation}
    {\mathbf{k}}_t = \mathbf{W}_k \mathbf{e}_t + \mathbf{b}_k,\,
    {\mathbf{v}}_t = \mathbf{W}_v \mathbf{e}_t + \mathbf{b}_v,
    \label{eq:kq}
\end{equation}
where $\mathbf{W}_k, \mathbf{W}_v \in \mathbb{R}^{M(n_{\text{max}}-n_{\text{min}}+1)d_h \times d}$ are learnt projections. 
The backbone hidden state $\mathbf{h}_t$ serves directly as the query, encoding what the current context expects at position $t$, while the key encodes what the n-gram memory offers.

We then compute a relevance score via scaled inner product. To stabilize training, we apply RMSNorm~\cite{zhang2019root} to both the key and query before taking the dot product:
\begin{equation}
    \alpha_t = \frac{\big\langle\, \text{RMSNorm}({\mathbf{k}}_t),\, \text{RMSNorm}({\mathbf{h}}_t) \,\big\rangle}{\sqrt{d}}.
    \label{eq:score}
\end{equation}
Since the sigmoid function suffers from vanishing gradients when scores are near zero, we apply a sign-preserving square root to compress the magnitude before activation:
\begin{equation}
    g_t = \sigma \big(\text{sign}(\alpha_t)\,\sqrt{|\alpha_t|}\,\big).
    \label{eq:gate}
\end{equation}
The scalar gate $g_t \in $ (0,1) then modulates the value vector to produce the gated n-gram contribution $\tilde{\mathbf{v}}_t = g_t \cdot \mathbf{v}_t$.
By conditioning on the backbone's hidden state, this mechanism contextualizes the static hash embeddings, suppressing irrelevant retrievals while amplifying informative substructure patterns. 
As we verify in \S\ref{ssec:analysis}, the gate learns to activate strongly at chemically meaningful positions (functional groups, reactive sites) and suppress at uninformative ones.

\paragraph{Short convolution.}
The gating mechanism operates independently at each position. However, chemical substructures are inherently \emph{compositional}: adjacent local patterns often combine into higher-order motifs. To enable interaction among neighboring n-gram representations and smooth the discrete hash-based signals, we apply a lightweight depthwise convolution~\citep{chollet2017xception} over the gated contributions:
\begin{equation}
\begin{aligned}
    \mathbf{z}_t &= \text{SiLU} \big( \text{DWConv}_{c}( \tilde{\mathbf{v}}_{t-c+1:t} ) \big), \\
    \hat{\mathbf{h}}_t &= \mathbf{h}_t + \tilde{\mathbf{v}}_t + \mathbf{z}_t,
\end{aligned}
\label{eq:shortconv}
\end{equation}
where $\text{DWConv}_{c}$ denotes a depthwise convolution with kernel size $c$
and $\text{SiLU}$~\citep{elfwing2017sigmoid} is the activation function. The residual connection adds both the gated signal $\tilde{\mathbf{v}}_t$ and the locally-composed signal $\mathbf{z}_t$ to the original hidden state. 

This \emph{gate-then-mix} design lets the gate select \emph{which} retrieved patterns are relevant, while the convolution determines \emph{how} neighbors compose (e.g., a carbonyl and an adjacent hydroxyl jointly signaling a carboxylic acid). The enriched representation $\hat{\mathbf{h}}_t$ is then passed to the next Transformer layer.

\subsection{Architecture-specific integration.}
We apply \model to two backbones: decoder-only as in \textbf{MolGPT}~\cite{bagal2021molgpt} and encoder-decoder as in \textbf{T5Chem}~\cite{lu2022unified}. For MolGPT, the module is inserted at selected decoder layers with causal convolution to preserve autoregression. 
For T5Chem, it is inserted at early encoder layers with bidirectional convolution, enriching source representations before cross-attention---so every decoder step benefits from substructure-aware encoding. 
Full configuration details are provided in Appendix~\ref{app:impl_details}.

\section{Experiments}
\label{sec:experiments}

We evaluate \model on three task settings covering unconditional molecule generation, forward reaction prediction, and single-step retrosynthesis, using two backbone architectures.

\subsection{Experimental Setup}
\label{ssec:setup}

\paragraph{Tasks, datasets, and metrics.}
Table~\ref{tab:datasets} summarizes the datasets used for the three evaluation tasks. 
(1) \textbf{Unconditional molecule generation} requires models to generate novel, valid, and realistic molecules without conditioning input. We use two standard benchmarks, MOSES~\cite{polykovskiy2020molecular} and GuacaMol~\cite{brown2019guacamol}, and report \textit{Validity}, \textit{Uniqueness}, \textit{Novelty}, \textit{Fr\'{e}chet ChemNet Distance} (FCD), \textit{Fragment similarity}, and \textit{Scaffold similarity} following the MOSES protocol.
(2) \textbf{Forward reaction prediction} predicts the major product from reactants and optionally reagents. We evaluate on USPTO-MIT~\cite{schneider2016s} with two input variants: \emph{separated}, where reactants and reagents are marked by a delimiter, and \emph{mixed}, where all species are concatenated without role distinction. 
(3) \textbf{Single-step retrosynthesis} predicts a valid set of reactants from a target product. We evaluate this task on USPTO-50k~\cite{schneider2016s}. 
For both reaction tasks, we report top-$k$ accuracy ($k \in \{1,2,5\}$) using beam search.

\paragraph{Backbone configurations.}
We use two Transformer backbones:   MolGPT~\cite{bagal2021molgpt} for GPT and T5Chem~\cite{lu2022unified} for T5. 
For each, we train models at multiple scales.
Detailed architecture specifications are provided in Appendix~\ref{app:impl_details}.

\paragraph{Implementation details.}
For both backbones, we use character-level tokenizers with a vocabulary size of 100. 
The models are trained from random initialization, without pretraining, using mixed-precision FP16. 
All experiments employ early stopping based on validation loss with a patience of 20 epochs. 
Reported accuracies are computed using \textit{checkpoint averaging} over the last 20 saved checkpoints. 
Detailed hyperparameter settings for each backbone are provided in Appendix~\ref{app:impl_details}.

\subsection{Unconditional Molecule Generation}
\label{ssec:main_generation}

\begin{table}[!ht]
    \centering
    \caption{Unconditional molecule generation results of GPT models on MOSES and GuacaMol.}
    \vspace{-0.1cm}
    \label{tab:uncond_mol_gen}
    \small
    \begin{threeparttable}
    \resizebox{\linewidth}{!}{
        \begin{tabular}{lcccccc}
        \toprule
        \multicolumn{7}{c}{\textit{MOSES}} \\
        \midrule
        \textbf{Model}
        & \textbf{Params}
        & \textbf{Val.$\uparrow$}
        & \textbf{Uniq.$\uparrow$}
        & \textbf{Nov.$\uparrow$}
        & \textbf{FCD$\downarrow$}
        & \textbf{Frag.$\uparrow$} \\
        \midrule
        MolGPT
        & 25.5M & 0.9951 & 0.9942 & 0.9783 & 0.0684 & \textbf{0.9992} \\
        \cellcolor{gray!15}\textbf{MolGram (GPT)}
        & \cellcolor{gray!15}27.2M
        & \cellcolor{gray!15}\textbf{0.9963} & \cellcolor{gray!15}\textbf{0.9947} & \cellcolor{gray!15}\textbf{0.9794} & \cellcolor{gray!15}\textbf{0.0673} & \cellcolor{gray!15}0.9991 \\
        \midrule
        MolGPT
        & 85.4M & 0.9960 & 0.9945 & 0.9786 & 0.0680 & 0.9992 \\
        \cellcolor{gray!15}\textbf{MolGram (GPT)}
        & \cellcolor{gray!15}87.5M
        & \cellcolor{gray!15}\textbf{0.9967} & \cellcolor{gray!15}\textbf{0.9954} & \cellcolor{gray!15}\textbf{0.9796} & \cellcolor{gray!15}\textbf{0.0669} & \cellcolor{gray!15}0.9992 \\

        \midrule
        \multicolumn{7}{c}{\textit{GuacaMol}} \\
        \midrule
        \textbf{Model}
        & \textbf{Params}
        & \textbf{Val.$\uparrow$}
        & \textbf{Uniq.$\uparrow$}
        & \textbf{Nov.$\uparrow$}
        & \textbf{FCD$^\dagger_{\mathrm{s}}\uparrow$}
        & \textbf{KL$\uparrow$} \\
        \midrule
        MolGPT
        & 25.5M & 0.9807 & 0.9914 & \textbf{0.9900} & 0.9085 & 0.9924 \\
        \cellcolor{gray!15}\textbf{MolGram (GPT)}
        & \cellcolor{gray!15}27.2M
        & \cellcolor{gray!15}\textbf{0.9821} & \cellcolor{gray!15}\textbf{0.9942} & \cellcolor{gray!15}0.9892 & \cellcolor{gray!15}\textbf{0.9101} & \cellcolor{gray!15}\textbf{0.9931} \\
        \midrule
        MolGPT
        & 85.4M & 0.9816 & 0.9922 & \textbf{0.9896} & 0.9096 & 0.9929 \\
        \cellcolor{gray!15}\textbf{MolGram (GPT)}
        & \cellcolor{gray!15}87.5M
        & \cellcolor{gray!15}\textbf{0.9829} & \cellcolor{gray!15}\textbf{0.9946} & \cellcolor{gray!15}0.9894 & \cellcolor{gray!15}\textbf{0.9112} & \cellcolor{gray!15}\textbf{0.9935} \\
        \bottomrule
        \end{tabular}
    }
    \begin{tablenotes}[flushleft]
    \footnotesize 
     \item[$^\dagger$] FCD$_{\mathrm{s}}$ is the normalized FCD score: $\exp$(-0.2 $\times$ FCD).
    \end{tablenotes}
    \end{threeparttable}
    \vspace{-0.5cm}
\end{table}

Table~\ref{tab:uncond_mol_gen} reports unconditional molecule generation results on MOSES and GuacaMol, where each model samples 30K molecules with temperature set to 1.0.
\model improves the MolGPT baselines on most metrics, with only negligible trade-offs in saturated scores.
On MOSES, \model consistently improves validity, uniqueness, novelty, and FCD at both scales, reducing FCD from 0.0684 to 0.0673 for the 25.5M model and from 0.0680 to 0.0669 for the 85.4M model.
On GuacaMol, \model improves validity, uniqueness, normalized FCD score, and KL divergence for both scales, with the largest gains in uniqueness.
Notably, MolGram (GPT) at 27.2M outperforms the larger MolGPT baseline at 85.4M on most metrics across both datasets, indicating that the gains are not simply due to increased parameter count. 
These results suggest that \model's explicit $n$-gram memory captures useful local chemical priors about recurring substructure combinations, improving molecular sequence generation beyond scale alone.

\subsection{Reaction Prediction}
\label{ssec:main_reaction}

\paragraph{Forward prediction.}

\begin{table*}[t]
    \centering
    \caption{Forward reaction prediction results on USPTO-MIT.}
    \vspace{-0.2cm}
    \label{tab:forward}
    \small
    \setlength{\tabcolsep}{4.8pt}
    \renewcommand{\arraystretch}{1.05}
    \begin{tabular}{clccccccccc}
        \toprule
        \multirow{2}{*}{\textbf{Backbone}}
        & \multirow{2}{*}{\textbf{Model}}
        & \multirow{2}{*}{\textbf{Params}}
        & \multicolumn{4}{c}{\textbf{Separated}}
        & \multicolumn{4}{c}{\textbf{Mixed}} \\
        \cmidrule(lr){4-7} \cmidrule(lr){8-11}
        &
        &
        & \textbf{Top-1}
        & \textbf{Top-2}
        & \textbf{Top-5}
        & \textbf{Valid.}
        & \textbf{Top-1}
        & \textbf{Top-2}
        & \textbf{Top-5}
        & \textbf{Valid.} \\
        \midrule

        \multirow{3}{*}{GPT}
        & MolGPT
        & 25.5M
        & 0.8806 & 0.9220 & 0.9441 & 0.9968
        & 0.8623 & 0.9069 & 0.9329 & 0.9972 \\
        & \cellcolor{gray!15}MolGram (GPT)
        & \cellcolor{gray!15}26.1M
        & \cellcolor{gray!15}0.8922
        & \cellcolor{gray!15}0.9302
        & \cellcolor{gray!15}0.9510
        & \cellcolor{gray!15}0.9975
        & \cellcolor{gray!15}0.8698
        & \cellcolor{gray!15}0.9127
        & \cellcolor{gray!15}0.9378
        & \cellcolor{gray!15}0.9972 \\
        & MolGPT
        & 85.4M
        & \textbf{0.8924} & \textbf{0.9309} & \textbf{0.9517} & \textbf{0.9976}
        & \textbf{0.8727} & \textbf{0.9156} & \textbf{0.9381} & \textbf{0.9981} \\

        \midrule
        \multirow{3}{*}{T5}
        & T5Chem
        & 14.7M
        & 0.8949 & 0.9371 & 0.9566 & 0.9974
        & 0.8742 & 0.9219 & 0.9461 & 0.9969 \\
        & \cellcolor{gray!15}MolGram (T5)
        & \cellcolor{gray!15}15.0M
        & \cellcolor{gray!15}\textbf{0.9043}
        & \cellcolor{gray!15}\textbf{0.9428}
        & \cellcolor{gray!15}\textbf{0.9650}
        & \cellcolor{gray!15}\textbf{0.9977}
        & \cellcolor{gray!15}\textbf{0.8872}
        & \cellcolor{gray!15}\textbf{0.9290}
        & \cellcolor{gray!15}\textbf{0.9535}
        & \cellcolor{gray!15}\textbf{0.9979} \\
        & T5Chem
        & 44.1M
        & 0.8963 & 0.9394 & 0.9613 & 0.9976
        & 0.8774 & 0.9243 & 0.9503 & 0.9977 \\

        \bottomrule
    \end{tabular}
\end{table*}

\begin{figure*}[t]
    \centering
    \resizebox{0.83\linewidth}{!}{
    \includegraphics{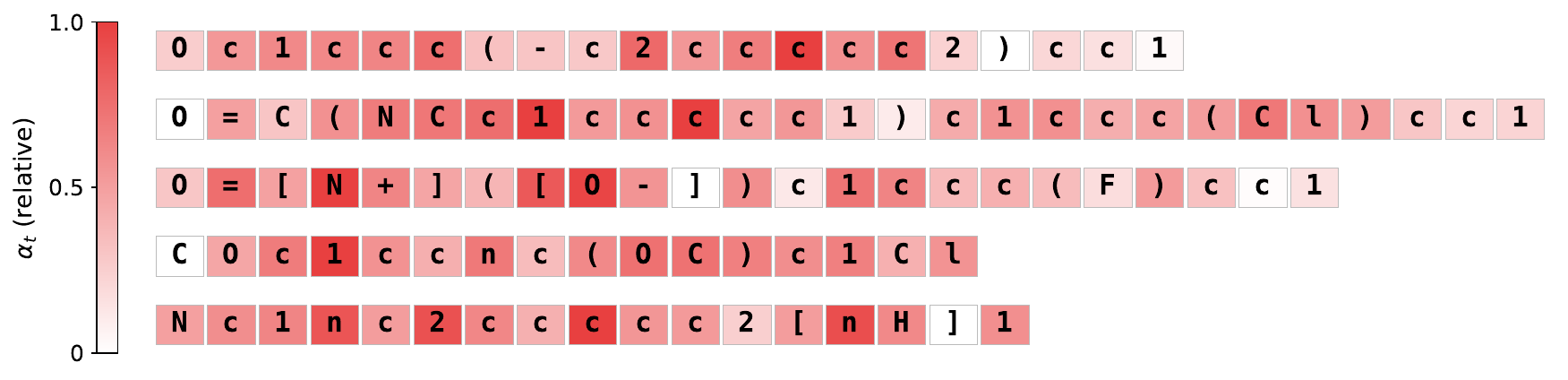}}
    \vspace{-0.2cm}
    \caption{Visualization of gate activations across SMILES positions of \model.}
    \vspace{-0.3cm}
    \label{fig:gating}
\end{figure*}

Table~\ref{tab:forward} further evaluates \model on forward reaction prediction.  Unlike unconditional generation, this task requires identifying the correct product from reactant and reagent patterns, making it a direct test of whether the learned local memory is useful for conditional chemical transformation.
Across both datasets and both backbone architectures, \model consistently improves accuracy at all top-$k$ levels. 
\model improves both GPT and T5 backbones across the separated and mixed settings, yielding relative top-1 accuracy gains of 1.3\% and 1.1\% in the \textit{separated} setting, and 0.9\% and 1.5\% in the \textit{mixed} setting. The improvements are also reflected at top-2 and top-5.
The comparison with larger baselines shows that these gains are not simply a consequence of parameter count: the 15.0M \model (T5) outperforms the 44.1M T5Chem model, and the 26.1M \model (GPT) matches or exceeds the 85.4M MolGPT baseline. Thus, the explicit $n$-gram memory appears to provide reaction-relevant local priors that transfer beyond molecule generation to conditional prediction.

\paragraph{Single-step retrosynthesis.}
\begin{table}[!ht]
    \centering
    \caption{Single-step retrosynthesis on USPTO-50k.}
    \vspace{-0.2cm}
    \label{tab:retro}
    \small
    \setlength{\tabcolsep}{4.5pt}
    \renewcommand{\arraystretch}{1.05}
    \resizebox{\linewidth}{!}{
        \begin{tabular}{lccccc}
            \toprule
            \textbf{Model}
            & \textbf{Params}
            & \textbf{Top-1}
            & \textbf{Top-2}
            & \textbf{Top-5}
            & \textbf{Valid.} \\
            \midrule

            MolGPT
            & 6.4M
            & 0.4085 & 0.5330 & 0.6491 & 0.9808 \\
            \cellcolor{gray!15}MolGram (GPT)
            & \cellcolor{gray!15}6.7M
            & \cellcolor{gray!15}\textbf{0.4209}
            & \cellcolor{gray!15}\textbf{0.5480}
            & \cellcolor{gray!15}\textbf{0.6673}
            & \cellcolor{gray!15}0.9808 \\
            MolGPT
            & 25.5M
            & 0.4168 & 0.5432 & 0.6635 & \textbf{0.9812} \\

            \midrule
            T5Chem
            & 14.7M
            & 0.4290 & 0.5650 & 0.6980 & 0.9792 \\
            \cellcolor{gray!15}MolGram (T5)
            & \cellcolor{gray!15}15.7M
            & \cellcolor{gray!15}\textbf{0.4460}
            & \cellcolor{gray!15}\textbf{0.5780}
            & \cellcolor{gray!15}\textbf{0.6990}
            & \cellcolor{gray!15}\textbf{0.9804} \\
            T5Chem
            & 44.1M
            & 0.4412 & 0.5756 & 0.6987 & 0.9800 \\

            \bottomrule
        \end{tabular}
    }
\end{table}
Table~\ref{tab:retro} reports single-step retrosynthesis results. 
\model improves top-1 accuracy by 3.0\% over MolGPT and by 4.0\% over T5Chem, with a 2.8\% top-5 gain for the GPT backbone, suggesting better ranking of plausible disconnections across the beam. 
Notably, the smaller MolGram variants also outperform their larger baselines: MolGram (GPT, 6.7M) exceeds MolGPT (25.5M), and MolGram (T5, 15.7M) exceeds T5Chem (44.1M). These results indicate that explicit $n$-gram memory is especially beneficial for retrosynthesis, where limited training data makes local functional-group and disconnection priors more valuable.

\subsection{\model gating visualization}
\label{ssec:analysis}

To validate whether \model captures meaningful chemical patterns, we visualize the per-token contribution ratio of the \model module on the forward reaction prediction task. Specifically, we measure $\|\mathbf{r}_t\|_2 / \|\mathbf{h}_t\|_2$ at each position, where $\mathbf{r}_t$ is the \model residual and $\mathbf{h}_t$ is the backbone hidden state. We visualize representative molecules from the test set in Figure~\ref{fig:gating}.

The \model module produces stronger contributions (darker red) at positions corresponding to meaningful local motifs: (1)~\textit{aromatic ring interiors}, where n-gram patterns like \texttt{ccc} and ring-opening digits encode ring regularity; (2)~\textit{functional group atoms}, including the nitro group \texttt{[N+]([O-])}, amide linkage \texttt{C(N}, and halide substitutions \texttt{(Cl)}; and (3)~\textit{heteroatom positions} such as pyridine nitrogen and ether oxygen. Conversely, structurally uninformative tokens---closing brackets, ring-closure endpoints, and isolated aliphatic carbons---consistently receive low activation.

\section{Conclusion}
\label{sec:conclusion}

In this work, we address the \textit{locality gap} in chemical language modeling by integrating a conditional $n$-gram memory into SMILES-based Transformers. 
By delegating local structural motifs to a dedicated hash-based lookup table, \model frees attention layers to focus on global structure and long-range reaction dependencies.
Across molecular generation, forward reaction prediction, and retrosynthesis, this model-side inductive bias consistently outperforms standard baselines and substantially larger models, with gating visualizations confirming selective activation at chemically meaningful substructures.
Compatible with both GPT and T5 backbones, \model offers a lightweight mechanism for injecting local structural awareness while preserving character-level simplicity.

\clearpage
\section*{Limitations}

While \model effectively bridges the locality gap in SMILES language models, several limitations remain. First, \model operates entirely on the surface text string representation rather than underlying molecular graph topologies. Because SMILES strings are highly sensitive to canonicalization protocols and starting-atom indexing, the same chemical functional group can map to completely different textual $n$-grams if the surrounding structural context shifts. Consequently, the model must rely on data augmentation (such as SMILES enumeration) to learn graph-invariant representations of these local memories. In addition, because the $n$-gram window size is strictly bounded ($n_{\text{min}} - n_{\text{max}}$), \model is fundamentally unequipped to capture non-local topological dependencies, such as distant ring-closure linkages or macrocyclic stereochemical relationships, which must still be resolved entirely by the backbone's global self-attention mechanism.



\bibliography{ref}

\begin{thebibliography}{26}
\providecommand{\natexlab}[1]{#1}

\bibitem[{Bagal et~al.(2021)Bagal, Aggarwal, Vinod, and Priyakumar}]{bagal2021molgpt}
Viraj Bagal, Rishal Aggarwal, PK~Vinod, and U~Deva Priyakumar. 2021.
\newblock Molgpt: molecular generation using a transformer-decoder model.
\newblock \emph{Journal of chemical information and modeling}, 62(9):2064--2076.

\bibitem[{Bjerrum(2017)}]{bjerrum2017smiles}
Esben~Jannik Bjerrum. 2017.
\newblock Smiles enumeration as data augmentation for neural network modeling of molecules.
\newblock \emph{arXiv preprint arXiv:1703.07076}.

\bibitem[{Brown et~al.(2019)Brown, Fiscato, Segler, and Vaucher}]{brown2019guacamol}
Nathan Brown, Marco Fiscato, Marwin~HS Segler, and Alain~C Vaucher. 2019.
\newblock Guacamol: benchmarking models for de novo molecular design.
\newblock \emph{Journal of chemical information and modeling}, 59(3):1096--1108.

\bibitem[{Cheng et~al.(2026)Cheng, Zeng, Dai, Chen, Wang, Xie, Huang, Yu, Hao, Li et~al.}]{cheng2026conditional}
Xin Cheng, Wangding Zeng, Damai Dai, Qinyu Chen, Bingxuan Wang, Zhenda Xie, Kezhao Huang, Xingkai Yu, Zhewen Hao, Yukun Li, and 1 others. 2026.
\newblock Conditional memory via scalable lookup: A new axis of sparsity for large language models.
\newblock \emph{arXiv preprint arXiv:2601.07372}.

\bibitem[{Chithrananda et~al.(2020)Chithrananda, Grand, and Ramsundar}]{chithrananda2020chemberta}
Seyone Chithrananda, Gabriel Grand, and Bharath Ramsundar. 2020.
\newblock Chemberta: large-scale self-supervised pretraining for molecular property prediction.
\newblock \emph{arXiv preprint arXiv:2010.09885}.

\bibitem[{Chollet(2017)}]{chollet2017xception}
Fran{\c{c}}ois Chollet. 2017.
\newblock Xception: Deep learning with depthwise separable convolutions.
\newblock In \emph{Proceedings of the IEEE conference on computer vision and pattern recognition}, pages 1251--1258.

\bibitem[{Deng et~al.(2025)Deng, Zhao, Sun, Chen, Wang, Xue, Li, Song, Hsieh, Hou et~al.}]{deng2025rsgpt}
Yafeng Deng, Xinda Zhao, Hanyu Sun, Yu~Chen, Xiaorui Wang, Xi~Xue, Liangning Li, Jianfei Song, Chang-Yu Hsieh, Tingjun Hou, and 1 others. 2025.
\newblock Rsgpt: a generative transformer model for retrosynthesis planning pre-trained on ten billion datapoints.
\newblock \emph{Nature communications}, 16(1):7012.

\bibitem[{Elfwing et~al.(2017)Elfwing, Uchibe, and Doya}]{elfwing2017sigmoid}
S~Elfwing, E~Uchibe, and K~Doya. 2017.
\newblock Sigmoid-weighted linear units for neural network function approximation in reinforcement learning. arxiv e-prints, art.
\newblock \emph{arXiv preprint arXiv:1702.03118}.

\bibitem[{Irwin et~al.(2022)Irwin, Dimitriadis, He, and Bjerrum}]{irwin2022chemformer}
Ross Irwin, Spyridon Dimitriadis, Jiazhen He, and Esben~Jannik Bjerrum. 2022.
\newblock Chemformer: a pre-trained transformer for computational chemistry.
\newblock \emph{Machine Learning: Science and Technology}, 3(1):015022.

\bibitem[{Lin et~al.(2025)Lin, Liu, Xiang, Zeng, and Zeng}]{lin2025enhancing}
Xuan Lin, Qingrui Liu, Hongxin Xiang, Daojian Zeng, and Xiangxiang Zeng. 2025.
\newblock Enhancing chemical reaction and retrosynthesis prediction with large language model and dual-task learning.
\newblock \emph{arXiv preprint arXiv:2505.02639}.

\bibitem[{Lu and Zhang(2022)}]{lu2022unified}
Jieyu Lu and Yingkai Zhang. 2022.
\newblock Unified deep learning model for multitask reaction predictions with explanation.
\newblock \emph{Journal of chemical information and modeling}, 62(6):1376--1387.

\bibitem[{Mastrolorito et~al.(2025)Mastrolorito, Ciriaco, Togo, Gambacorta, Trisciuzzi, Altomare, Amoroso, Grisoni, and Nicolotti}]{mastrolorito2025fragsmiles}
Fabrizio Mastrolorito, Fulvio Ciriaco, Maria~Vittoria Togo, Nicola Gambacorta, Daniela Trisciuzzi, Cosimo~Damiano Altomare, Nicola Amoroso, Francesca Grisoni, and Orazio Nicolotti. 2025.
\newblock fragsmiles as a chemical string notation for advanced fragment and chirality representation.
\newblock \emph{Communications Chemistry}, 8(1):26.

\bibitem[{Noutahi et~al.(2024)Noutahi, Gabellini, Craig, Lim, and Tossou}]{noutahi2024gotta}
Emmanuel Noutahi, Cristian Gabellini, Michael Craig, Jonathan~SC Lim, and Prudencio Tossou. 2024.
\newblock Gotta be safe: a new framework for molecular design.
\newblock \emph{Digital Discovery}, 3(4):796--804.

\bibitem[{Polykovskiy et~al.(2020)Polykovskiy, Zhebrak, Sanchez-Lengeling, Golovanov, Tatanov, Belyaev, Kurbanov, Artamonov, Aladinskiy, Veselov et~al.}]{polykovskiy2020molecular}
Daniil Polykovskiy, Alexander Zhebrak, Benjamin Sanchez-Lengeling, Sergey Golovanov, Oktai Tatanov, Stanislav Belyaev, Rauf Kurbanov, Aleksey Artamonov, Vladimir Aladinskiy, Mark Veselov, and 1 others. 2020.
\newblock Molecular sets (moses): a benchmarking platform for molecular generation models.
\newblock \emph{Frontiers in pharmacology}, 11:565644.

\bibitem[{Radford et~al.(2019)Radford, Wu, Child, Luan, Amodei, Sutskever et~al.}]{radford2019language}
Alec Radford, Jeffrey Wu, Rewon Child, David Luan, Dario Amodei, Ilya Sutskever, and 1 others. 2019.
\newblock Language models are unsupervised multitask learners.
\newblock \emph{OpenAI blog}, 1(8):9.

\bibitem[{Raffel et~al.(2019)Raffel, Shazeer, Roberts, Lee, Narang, Matena, Zhou, Li, and Liu}]{raffel2019exploring}
C~Raffel, Noam Shazeer, A~Roberts, K~Lee, S~Narang, M~Matena, Y~Zhou, W~Li, and PJ~Liu. 2019.
\newblock Exploring the limits of transfer learning with a unified text-to-text transformer. arxiv preprint arxiv: 191010683.
\newblock \emph{Published online}.

\bibitem[{Ross et~al.(2022)Ross, Belgodere, Chenthamarakshan, Padhi, Mroueh, and Das}]{ross2022large}
Jerret Ross, Brian Belgodere, Vijil Chenthamarakshan, Inkit Padhi, Youssef Mroueh, and Payel Das. 2022.
\newblock Large-scale chemical language representations capture molecular structure and properties.
\newblock \emph{Nature Machine Intelligence}, 4(12):1256--1264.

\bibitem[{Sagawa and Kojima(2023)}]{sagawa2023reactiont5}
Tatsuya Sagawa and Ryosuke Kojima. 2023.
\newblock Reaction{T5}: a large-scale pre-trained model towards application of limited reaction data.
\newblock \emph{arXiv preprint arXiv:2311.06708}.

\bibitem[{Schneider et~al.(2016)Schneider, Stiefl, and Landrum}]{schneider2016s}
Nadine Schneider, Nikolaus Stiefl, and Gregory~A Landrum. 2016.
\newblock What’s what: The (nearly) definitive guide to reaction role assignment.
\newblock \emph{Journal of chemical information and modeling}, 56(12):2336--2346.

\bibitem[{Schwaller et~al.(2019)Schwaller, Laino, Gaudin, Bolgar, Hunter, Bekas, and Lee}]{schwaller2019molecular}
Philippe Schwaller, Teodoro Laino, Th{\'e}ophile Gaudin, Peter Bolgar, Christopher~A Hunter, Costas Bekas, and Alpha~A Lee. 2019.
\newblock Molecular transformer: a model for uncertainty-calibrated chemical reaction prediction.
\newblock \emph{ACS central science}, 5(9):1572--1583.

\bibitem[{Sennrich et~al.(2016)Sennrich, Haddow, and Birch}]{sennrich2016neural}
Rico Sennrich, Barry Haddow, and Alexandra Birch. 2016.
\newblock Neural machine translation of rare words with subword units.
\newblock In \emph{Proceedings of the 54th annual meeting of the association for computational linguistics (volume 1: long papers)}, pages 1715--1725.

\bibitem[{Tetko et~al.(2020)Tetko, Karpov, Van~Deursen, and Godin}]{tetko2020state}
Igor~V Tetko, Pavel Karpov, Ruud Van~Deursen, and Guillaume Godin. 2020.
\newblock State-of-the-art augmented nlp transformer models for direct and single-step retrosynthesis.
\newblock \emph{Nature communications}, 11(1):5575.

\bibitem[{Weininger(1988)}]{weininger1988smiles}
David Weininger. 1988.
\newblock Smiles, a chemical language and information system. 1. introduction to methodology and encoding rules.
\newblock \emph{Journal of chemical information and computer sciences}, 28(1):31--36.

\bibitem[{Xu et~al.(2026)Xu, Feng, Chen, Liu, Deng, Ding, Long, Shuai, Li, Liu et~al.}]{xu2026beyond}
Huinan Xu, Xuyang Feng, Junhong Chen, Junchen Liu, Kaiwen Deng, Kai Ding, Shengning Long, Jiaxue Shuai, Zhaorong Li, Shiping Liu, and 1 others. 2026.
\newblock Beyond conditional computation: Retrieval-augmented genomic foundation models with gengram.
\newblock \emph{arXiv preprint arXiv:2601.22203}.

\bibitem[{Zhang and Sennrich(2019)}]{zhang2019root}
Biao Zhang and Rico Sennrich. 2019.
\newblock Root mean square layer normalization.
\newblock \emph{Advances in neural information processing systems}, 32.

\bibitem[{Zhang et~al.(2025)Zhang, Li, Chen, Zhao, Lin, Zhu, Chen, Chen, and Yu}]{zhang2025reasoning}
Situo Zhang, Hanqi Li, Lu~Chen, Zihan Zhao, Xuanze Lin, Zichen Zhu, Bo~Chen, Xin Chen, and Kai Yu. 2025.
\newblock Reasoning-driven retrosynthesis prediction with large language models via reinforcement learning.
\newblock \emph{arXiv preprint arXiv:2507.17448}.

\end{thebibliography}

\appendix

\appendix

\section{LLM USAGE}
LLM was used as a writing assistance tool to improve the linguistic quality and readability of this manuscript. Specifically, we employed it for tasks such as sentence rephrasing, grammar correction, and enhancing the clarity and flow of the text presentation.

The authors take full responsibility for all content in this manuscript, including any LLM-assisted text, and have verified that all LLM usage adheres to ethical guidelines and does not contribute to plagiarism or scientific misconduct.

\section{Dataset Details}
\label{app:datasets}

\noindent\textbf{USPTO-MIT.} Derived from the United States Patent and Trademark Office database~\citep{schneider2016s}, this dataset contains 479,035 reactions covering 10 reaction types (4409,035 train / 30,000 validation / 40,000 test). The \emph{separated} variant delimits reactants and reagents with a special token; the \emph{mixed} variant concatenates all input species.

\vspace{8pt}
\noindent\textbf{USPTO-50k.} A subset of 50,037 reactions classified into 10 types~\citep{schneider2016s}, widely used for retrosynthesis benchmarking. We use the canonical split of 40,029 / 5,004 / 5,004.

\vspace{8pt}
\noindent\textbf{MOSES.} A collection of 1,936,963 molecules from ZINC Clean Leads~\citep{polykovskiy2020molecular}, filtered by molecular weight ($\leq$500), LogP ($\leq$5), and other drug-likeness criteria. We use the standard split.

\vspace{8pt}
\noindent\textbf{GuacaMol.} This benchmark contains 1,591,378 molecules from ChEMBL~\citep{brown2019guacamol}, filtered for quality, and provides standard splits and evaluation scripts.

\section{Implementation Details}
\label{app:impl_details}

Table~\ref{tab:impl_gpt2} and Table~\ref{tab:impl_t5} provide the full training configurations for GPT and T5 backbones, respectively.

\paragraph{Larger model configurations.}
For parameter fairness experiments, we scale both backbones without \model:
\begin{itemize}[leftmargin=*, itemsep=2pt]
    \item \textbf{GPT-25.5M:} 8 layers, 512 dim, 8 heads.
    \item \textbf{GPT-85.4M:} 12 layers, 768 dim, 12 heads.
    \item \textbf{T5-14.7M:} 4 decoder layers, 4 encoder layers, 8 heads, 256 dim, FFN 2048.
    \item \textbf{T5-44.1M:} 6 decoder layers, 6 encoder layers, 8 heads, 512 dim, FFN 2048.
\end{itemize}
All larger models use the same training recipe (LR, schedule, batch size, early stopping) as their smaller counterparts to ensure a fair comparison.

\paragraph{Character-level tokenizer.}
Our tokenizer uses a simpler character-level scheme with 100 tokens, following prior work~\citep{lu2022unified}.

\begin{table*}[!th]
    \centering
    \caption{Dataset statistics across all evaluation tasks.}
    \label{tab:datasets}
    \resizebox{0.6\linewidth}{!}{
        \begin{tabular}{lcccl}
            \toprule
            \textbf{Dataset} & \textbf{\#Train} & \textbf{\#Valid} & \textbf{\#Test} & \textbf{Task} \\
            \midrule
            MOSES & 1,505,429 & 79,234 & 176,074 & Molecule generation \\
            GuacaMol & 1,145,793 & 63,655 & 63,656 & Molecule generation \\
            USPTO-MIT & 409,035 & 30,000 & 40,000 & Forward prediction \\
            USPTO-50k & 40,029 & 5,004 & 5,004 & Retrosynthesis \\
            \bottomrule
        \end{tabular}
    }
\end{table*}

\begin{table}[!t]
    \centering
    \caption{Training configuration for GPT backbone.}
    \label{tab:impl_gpt2}
    \small
    \resizebox{\columnwidth}{!}{
    \begin{tabular}{ll}
        \toprule
        \textbf{Hyperparameter} & \textbf{Value} \\
        \midrule
        \multicolumn{2}{l}{\emph{Tokenizer}} \\
        \quad Type & character-level \\
        \quad Vocab size & 100 \\
        \midrule
        \multicolumn{2}{l}{\emph{Training}} \\
        \quad Learning rate & $1 \times 10^{-3}$ \\
        \quad LR schedule & Inverse square root \\
        \quad Warmup steps & 8,000 \\
        \quad Max training steps & 80,000 \\
        \quad Batch size & 256 \\
        \quad Loss & Target-only \\
        \quad Early stopping patience & 20 (on val loss) \\
        \quad Precision & FP16 \\
        \midrule
        \multicolumn{2}{l}{\emph{Engram module (when enabled)}} \\
        \quad Injection layers & [1, 4] \\
        \quad N-gram range & 2--3 / 3--4 / 4--6 \\
        \quad Vocab multiplier & 5 \\
        \quad Kernel size & 4 \\
        \quad Heads per n-gram & 4 \\
        \quad Embed dim per head & 64 \\
        \quad Engram dropout & 0.1 / 0.15 \\
        \midrule
        \multicolumn{2}{l}{\emph{Evaluation}} \\
        \quad Beam size & 10 \\
        \quad Checkpoint averaging & Last 20 checkpoints \\
        \bottomrule
    \end{tabular}
    }
\end{table}

\begin{table}[!t]
    \centering
    \caption{Training configuration for T5 backbone.}
    \label{tab:impl_t5}
    \small
    \resizebox{\columnwidth}{!}{
    \begin{tabular}{ll}
        \toprule
        \textbf{Hyperparameter} & \textbf{Value} \\
        \midrule
        \multicolumn{2}{l}{\emph{Tokenizer}} \\
        \quad Type & character-level \\
        \quad Vocab size & 100 \\
        \midrule
        \multicolumn{2}{l}{\emph{Training}} \\
        \quad Learning rate & $1 \times 10^{-3}$ \\
        \quad LR schedule & Linear \\
        \quad Warmup steps & 500  \\
        \quad Max epochs & 50 \\
        \quad Batch size & 256 \\
        \quad Loss & Standard seq2seq \\
        \quad Early stopping patience & 20 (on val loss) \\
        \quad Precision & FP16 \\
        \midrule
        \multicolumn{2}{l}{\emph{Engram module (when enabled)}} \\
        \quad Injection layers & [0, 2] / [0, 1, 2, 3] \\
        \quad N-gram range & 2--3 / 2--4 \\
        \quad Vocab multiplier & 5 \\
        \quad Kernel size & 4 \\
        \quad Heads per n-gram & 4 \\
        \quad Embed dim per head & 64 \\
        \quad Engram dropout & 0.1 \\
        \midrule
        \multicolumn{2}{l}{\emph{Evaluation}} \\
        \quad Beam size & 10 \\
        \quad Checkpoint averaging & Last 20 checkpoints \\
        \bottomrule
    \end{tabular}
    }
\end{table}

\end{document}